\documentclass[11pt,a4paper]{article}
\usepackage{authblk}
\usepackage[hyperref]{acl2019}
\usepackage{times}
\usepackage{latexsym}
\usepackage{multirow}

\usepackage{url}
\usepackage{mathtools}
\usepackage{amssymb}
\usepackage{amsmath}
\usepackage{booktabs}
\usepackage{subcaption}
\usepackage{multirow}
\usepackage{todonotes}
\usepackage{algorithm2e}
\usepackage{paralist}
\usepackage{xspace}

\newcommand{\clqg}{\textsc{CLQG}}

\DeclareMathOperator*{\argmax}{arg\,max}
\aclfinalcopy 

\title{Cross-Lingual Training for Automatic Question Generation}

\author[1,2]{Vishwajeet Kumar}
\author[2]{Nitish Joshi}
\author[2]{Arijit Mukherjee}
\author[2]{Ganesh Ramakrishnan}
\author[2]{Preethi Jyothi}
\affil[1]{IITB-Monash Research Academy, Mumbai, India}
\affil[2]{IIT Bombay, Mumbai, India}
\affil[ ]{\texttt{\{vishwajeet, nitishj, ganesh, pjyothi\}@cse.iitb.ac.in}}
\affil[ ]{\texttt{\{arijitmukh007\}@gmail.com}}
\date{}

\begin{document}
\maketitle
\begin{abstract}
Automatic question generation (QG) is a challenging problem in natural language understanding. QG systems are typically built assuming access to a large number of training instances where each instance is a question and its corresponding answer. For a new language, such training instances are hard to obtain making the QG problem even more challenging. Using this as our motivation, we study the reuse of an available large QG dataset in a secondary language (e.g. English) to learn a QG model for a primary language (e.g. Hindi) of interest. For the primary language, we assume access to a large amount of monolingual text but only a small QG dataset. We propose a cross-lingual QG model which uses the following training regime: (i) Unsupervised pretraining of language models in both primary and secondary languages and (ii) joint supervised training for QG in both languages. We demonstrate the efficacy of our proposed approach using two different primary languages, Hindi and Chinese. 
We also create and release a new question answering dataset for Hindi consisting of 6555 sentences.

\end{abstract}

\section{Introduction}

Automatic question generation from text is an important yet challenging problem especially when there is limited training data (i.e., pairs of sentences and corresponding questions). Standard sequence to sequence models for automatic question generation have been shown to perform reasonably well for languages like English, for which hundreds of thousands of training instances are available. However, training sets of this size are not available for most languages.
Manually curating a dataset of comparable size for a new language will be tedious and expensive. Thus, it would be desirable to  leverage existing question answering datasets to help build QG models for a new language. This is the overarching idea that motivates this work. In this paper, we present a cross-lingual model for leveraging a large question answering dataset in a secondary language (such as English) to train models for QG in a primary language (such as Hindi) with a significantly smaller question answering dataset. 
\begin{figure}[t!]
\centering
\includegraphics[width=0.45\textwidth]{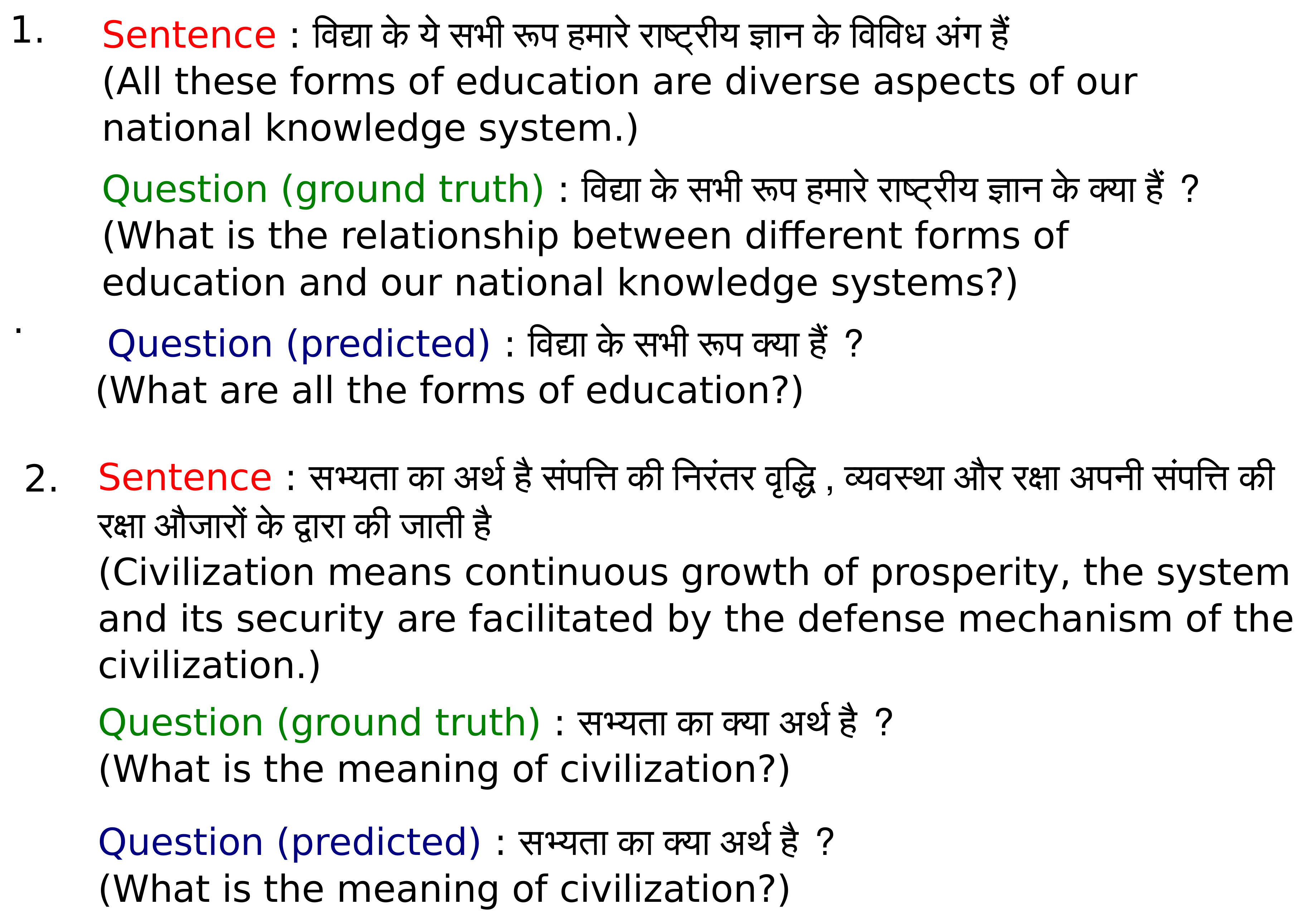}
\caption{Automatic QG from Hindi text.}
\label{fig:example-1}
\end{figure}

\par We chose Hindi to be one of our primary languages.
There is no established dataset available for Hindi that can be used to build question answering or question generation systems, making it an appropriate choice as a primary language. We create a new question answering dataset for Hindi (named \textbf{HiQuAD}):~\url{https://www.cse.iitb.ac.in/~ganesh/HiQuAD/clqg/}. Figure~\ref{fig:example-1} shows two examples of sentence-question pairs from \textbf{HiQuAD} along with the questions predicted by our best model. We also experimented with Chinese as a primary language. This choice was informed by our desire to use a language that was very different from Hindi. We use the same secondary language -- English -- with both choices of our primary language.

\par Drawing inspiration from recent work on unsupervised neural machine translation \cite{artetxe2017unsupervised,Yang2018UnsupervisedNM}, we propose a cross-lingual model to leverage resources available in a secondary language while learning to automatically generate questions from a primary language. We first train models for alignment between the primary and secondary languages in an unsupervised manner using monolingual text in both languages. We then use the relatively larger QG dataset in a secondary language to improve QG on the primary language.
Our main contributions can be summarized as follows:
\begin{asparaitem}
    \item We present a cross-lingual model that effectively exploits resources in a secondary language to improve QG for a primary language.  
    \item We demonstrate the value of cross-lingual training for QG using two primary languages, Hindi and Chinese. 
    \item We create a new question answering dataset for Hindi, \textbf{HiQuAD}.
\end{asparaitem}

\section{Related Work}
\label{rel_works}
Prior work in QG from text can be classified into two broad categories.

\paragraph{Rule-based:} Rule-based approaches~\cite{heilman2011automatic} mainly rely on manually curated rules for transforming a declarative sentence into an interrogative sentence. The quality of the  questions generated using rule-based systems highly depends on the quality of the handcrafted rules. Manually curating a large number of rules for a new language is a tedious and challenging task. More recently,~\citet{zheng2018novel} propose a template-based technique to construct questions from Chinese text, where they rank generated questions using a neural model and select the top-ranked question as the final output.

\paragraph{Neural Network Based:} Neural network based approaches do not rely on hand-crafted rules, but instead use an encoder-decoder architecture which can be trained in an end-to-end fashion to automatically generate questions from text.
Several neural network based approaches ~\cite{du2017learning,kumar2018automating,kumar2018framework} have been proposed for automatic question generation from text. ~\citet{du2017learning} propose a sequence to sequence model for automatic question generation from English text. ~\citet{kumar2018automating} use a rich set of linguistic features and encode pivotal answers predicted using a pointer network based model to automatically generate a question for the encoded answer. All existing models optimize a cross-entropy based loss function, that suffers from exposure bias~\cite{ranzato2015sequence}. Further, existing methods do not directly address the problem of handling important rare words and word repetition in QG. ~\citet{kumar2018framework} propose a reinforcement learning based framework which addresses the problem of exposure bias, word repetition and rare words. ~\citet{tang2017question} and ~\citet{wang2017joint} propose a joint model to address QG and the question answering problem together.

All prior work on QG assumed access to a sufficiently large number of training instances for a language. We relax this assumption in our work as we only have access to a small question answering dataset in the primary language. We show how we can improve QG performance on the primary language by leveraging a larger question answering dataset in a secondary language.  (Similarly in spirit, cross-lingual transfer learning based approaches have been recently proposed for other NLP tasks such as machine translation~\cite{schuster2018cross,lample2019cross}.) 

\section{Our Approach}
\label{sec:main_model}
We propose a shared encoder-decoder architecture that is trained in two phases. The first, is an \textbf{unsupervised pretraining} phase, consisting of denoising autoencoding and back-translation. This pretraining phase only requires sentences in both the primary and secondary languages. This is followed by a \textbf{supervised question generation} training phase that uses sentence-question pairs in both languages to fine-tune the pretrained weights. 
\begin{figure}
    \centering
    \includegraphics[width=\linewidth]{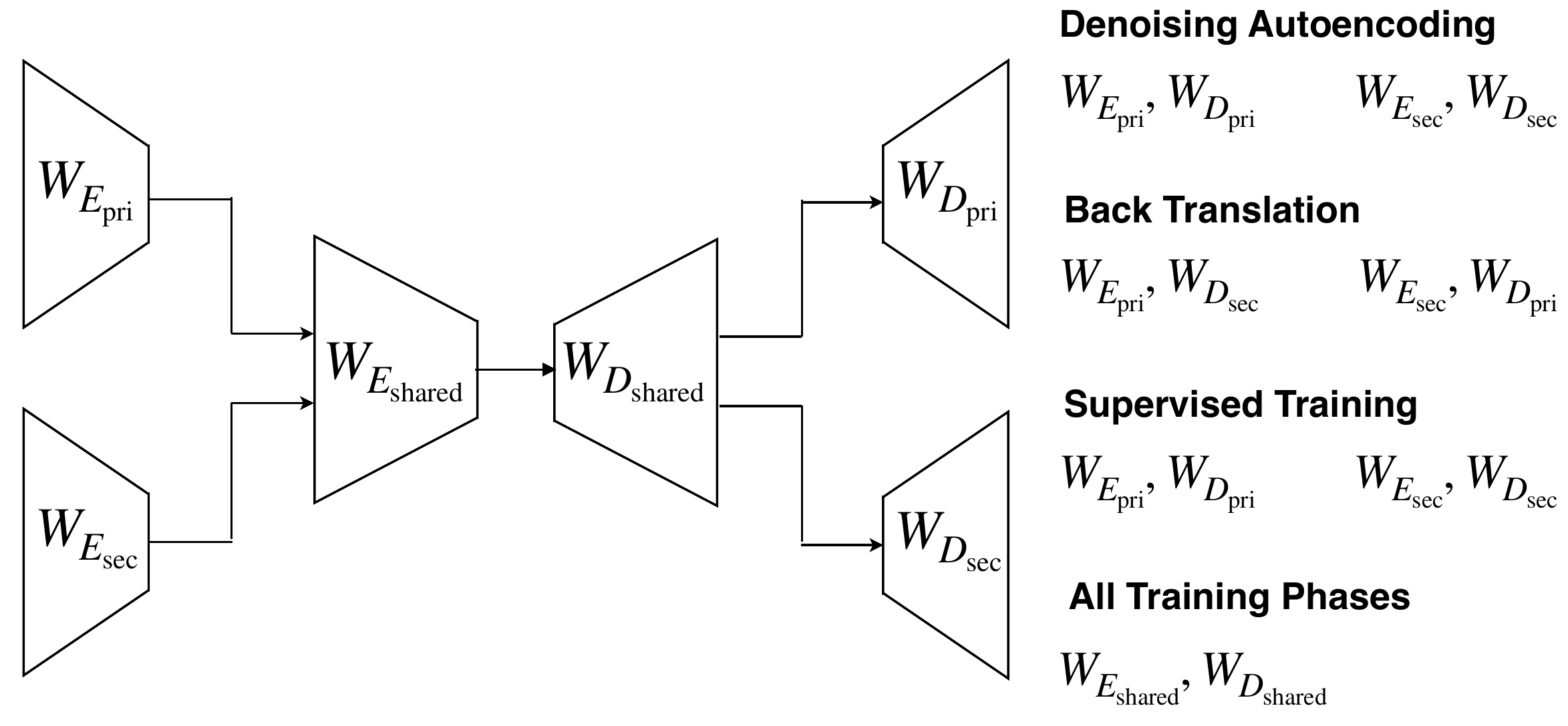}
    \caption{Schematic diagram of our cross-lingual QG system. $W_{E_{\mathrm{pri}}}$ and $W_{E_{\mathrm{sec}}}$ refer to parameters of the encoder layers specific to the primary and secondary languages; $W_{D_{\mathrm{pri}}}$ and $W_{D_{\mathrm{sec}}}$ are the weights of the corresponding decoder layers. $W_{E_{\mathrm{shared}}}$ and $W_{D_{\mathrm{shared}}}$ refer to weights of the encoder and decoder layers shared across both languages, respectively. Weights updated in each training phase are explicitly listed.}
    \label{fig:arch}
\end{figure}

\begin{algorithm}[t]
\SetAlgoLined
      \nl \textbf{Unsupervised Pretraining}\\
      \While{not converged}{
      \small
          \nl Train autoencoder to generate sentence \(x_p\) from noisy sentence \(\Tilde{x_p}\) in primary language and similarly \(x_s\)  from \(\Tilde{x_s}\) in the secondary language.\\
          \nl Back Translation: Generate sentences \(x_p^{'}\) and \(x_s{'}\) in primary and secondary \nl languages from \(x_s\) and \(x_p\) respectively, using the current translation model.\\
          \nl Train a new translation model using \(x_p^{'}\) and \(x_s{'}\) where \(x_s\) and \(x_p\) are used for supervision, respectively.
    
    }
    \nl \textbf{Supervised Question Generation}\\
    \nl {\small Initialize with pretrained weights}\\
    \While{not converged}{
    \small
    \nl Train sequence to sequence models for question generation in both the primary and secondary languages.
    }
  \caption{Cross-lingual Training Algorithm for QG}
  \label{alg:training}
\end{algorithm}

In Algorithm 1, we outline our training procedure and Figure~\ref{fig:arch} illustrates the overall architecture of our QG system. Our cross-lingual QG model consists of two encoders and two decoders specific to each language. We also enforce shared layers in both the encoder and the decoder whose weights are updated using data in both languages. (This weight sharing is discussed in more detail in Section~\ref{ssec:more}.)
For the encoder and decoder layers, we use the newly released Transformer~\cite{vaswani2017attention} model that has shown great success compared to recurrent neural network-based models in neural machine translation. Encoders and decoders consist of a stack of four identical layers, of which two layers are independently trained and two are trained in a shared manner. Each layer of the transformer consists of a multi-headed self-attention model followed by a  position-wise fully connected feed-forward network. 
\subsection{Unsupervised Pretraining}
We use monolingual corpora available in the primary (Hindi/Chinese) and secondary (English) languages for unsupervised pretraining. 
Similar to~\citet{artetxe2017unsupervised}, we use denoising autoencoders along with back-translation (described in Section \ref{back_tr}) for pretraining the language models in both the primary and secondary languages. Specifically, we first train the model to reconstruct their inputs, which will expose the model to the grammar and vocabulary specific to each language while enforcing a shared latent-space with the help of the shared encoder and decoder layers. To prevent the model from simply learning to copy every word, we randomly permute the word order in the input sentences so that the model learns meaningful structure in the language. If $x_p$ denotes the true input sentence to be generated from the sentence with permuted word order $\Tilde{x_p}$ for the primary language, then during each pass of the autoencoder training we update the weights $W_{E_{\mathrm{pri}}}$, $W_{E_{\mathrm{shared}}}$, $W_{D_{\mathrm{shared}}}$ and $W_{D_{\mathrm{pri}}}$. For the secondary language, we analogously update $W_{E_{\mathrm{sec}}}$, $W_{D_{\mathrm{sec}}}$ and the weights in the shared layers as shown in Figure~\ref{fig:arch}.

\subsubsection{Back translation}
\label{back_tr}
In addition to denoising autoencoders, we utilize back-translation~\cite{Sennrich2016ImprovingNM}. This further aids in enforcing the shared latent space assumption by generating a pseudo-parallel corpus~\cite{imankulova2017improving}.\footnote{A pseudo-parallel corpus consists of pairs of translated sentences using the current state of the model along with the original sentences.} Back translation has been demonstrated to be very important for unsupervised NMT~\cite{Yang2018UnsupervisedNM, Lample2018PhraseBasedN}. Given a sentence in the secondary language $x_s$, we generate a translated sentence in the primary language, $\Tilde{x_p}$. We then use the translated sentence $\Tilde{x_p}$ to generate the original $x_s$ back, while updating the weights $W_{E_{\mathrm{sec}}}$, $W_{E_{\mathrm{shared}}}$, $W_{D_{\mathrm{shared}}}$ and $W_{D_{\mathrm{pri}}}$ as shown in Figure~\ref{fig:arch}. Note that we utilize denoising autoencoding and back-translation  for both languages in each step of training.

\subsection{Supervised Question Generation}
We formulate the QG problem as a sequence to sequence modeling task where the input is a sentence and the output is a semantically consistent, syntactically correct and relevant question in the same language that corresponds to the sentence. Each encoder receives a sentence $\mathbf{x}$ (from the corresponding language) as input and the decoder generates a question $\mathbf{\Bar{y}}$
such that $\mathbf{\Bar{y}} = \argmax_{y}P(\mathbf{y}|\mathbf{x})$, and  $P(\mathbf{y}|\mathbf{x}) = \displaystyle \prod_{t=1}^{|y|} P(y_t | \mathbf{x}, y_{<t})$, where probability of each subword $y_t$ is predicted conditioned on all the subwords generated previously $y_{<t}$ and the input sentence $\mathbf{x}$.
We initialize the encoder and decoder weights using unsupervised pretraining and fine-tune these weights further during the supervised QG model training. Specifically, in each step of training, we update the weights $W_{E_{\mathrm{sec}}}$, $W_{E_{\mathrm{shared}}}$, $W_{D_{\mathrm{shared}}}$ and $W_{D_{\mathrm{sec}}}$ using QG data in the secondary language and $W_{E_{\mathrm{pri}}}$, $W_{E_{\mathrm{shared}}}$, $W_{D_{\mathrm{shared}}}$ and $W_{D_{\mathrm{pri}}}$ using QG data in the primary language. 
\subsection{More Architectural Details}
\label{ssec:more}

\noindent We make three important design choices:
\begin{enumerate}
\item \textbf{Use of positional masks:} ~\citet{shen2018disan} point out that transformers are not capable of capturing within the attention, information about order of the sequence. Following~\citet{shen2018disan}, we enable our encoders to use directional self attention so that temporal information is preserved. We use positional encodings which are essentially sine and cosine functions of different frequencies. More formally, positional encoding (PE) is defined as:
\begin{align}
    & PE_{(\text{pos},2i)} = \sin{\left( \frac{\text{pos}}{m^{\frac{2i}{d_{\text{model}}}}} \right)}\\
    & PE_{(\text{pos},2i+1)} = \cos{\left( \frac{\text{pos}}{m^{\frac{2i}{d_{\text{model}}}}} \right)}
\end{align}
where $m$ is a hyper-parameter, $\text{pos}$ is the position, $d_{\text{model}}$ is the dimensionality of the transformer and $i$ is the dimension. Following ~\citet{vaswani2017attention}, we set $m$ to 10000 in all our experiments. Directional self attention uses positional masks to inject temporal order information. Based on ~\citet{shen2018disan}, we define a forward positional mask ($M^f$) and a backward positional mask ($M^b$),
 \begin{align}
    &M_{ij}^{f} = \begin{cases}
                0, & i < j.\\
                -\infty, & \text{otherwise}.
                \end{cases} \nonumber \\
    &M_{ij}^{b} = \begin{cases}
                0, & i > j.\\
                -\infty, & \text{otherwise}.
                \end{cases} \nonumber
 \end{align}
that processes the sequence in the forward and backward direction, respectively.
\item \textbf{Weight sharing:} Based on the assumption that sentences and questions in two languages are similar in some latent space, in order to get a shared language independent representation, we share the last few layers of the encoder and the first few layers of the decoder~\cite{Yang2018UnsupervisedNM}. Unlike~\citet{artetxe2017unsupervised, Lample2018PhraseBasedN}, we do not share the encoder completely across the two languages, thus allowing the encoder layers private to each language to capture language-specific information. We found this to be useful in our experiments.
\item \textbf{Subword embeddings}: We represent data using BPE (Byte Pair Encoding)~\cite{gage1994new} embeddings.  We use BPE embeddings for both unsupervised pretraining as well as the supervised QG training phase. This allows for more fine-grained control over input embeddings compared to word-level embeddings~\cite{Sennrich2016NeuralMT}. This also has the advantage of maintaining a relatively smaller vocabulary size.\footnote{Using word embeddings across pretraining and the main QG task makes the vocabulary very large, thus leading to large memory issues.} 
\end{enumerate}

\section{Experimental Setup}
\label{exp}
We first describe all the datasets we used in our experiments, starting with a detailed description of our new Hindi question answering dataset, ``\textbf{HiQuAD}". We will then describe various implementation-specific details relevant to training our models. We conclude this section with a description of our evaluation methods.
\subsection{Datasets}
\subsubsection{HiQuAD}
\label{hiprep}
HiQuAD (Hindi Question Answering dataset) is a new question answering dataset in Hindi that we developed for this work. This dataset contains 6555 question-answer pairs from 1334 paragraphs in a series of books called Dharampal Books. 
\footnote{HiQuAD can be downloaded from:~\url{https://www.cse.iitb.ac.in/~ganesh/HiQuAD/clqg/}}

Similar to SQuAD ~\cite{rajpurkar2016squad}, an English question answering dataset that we describe further in Section~\ref{sec:datasets}, HiQuAD also consists of a paragraph, a list of questions answerable from the paragraph and answers to those questions. To construct sentence-question pairs, for a given question, we identified the first word of the answer in the paragraph and extracted the corresponding sentence to be paired along with the question. We curated a total of 6555 sentence-question pairs.
 
We tokenize the sentence-question pairs to remove any extra white spaces. For our experiments, we randomly split the HiQuAD dataset into train, development and test sets as shown in Table~\ref{tab:hidetails}. All model hyperparameters are optimized using the development set and all results are reported on the test set. 

\begin{table}[t!]
    \centering
    \begin{tabular}{c|c}
    \toprule
        \#pairs (Train set)  & 4000  \\
        \#pairs (Dev set) & 1300 \\
        \#pairs (Test set) & 1255 \\ 
    \midrule
    Text: avg tokens &  28.64\\
    Question: avg tokens & 14.13\\
    \bottomrule
    \end{tabular}
    \caption{HiQuAD dataset details}
    \label{tab:hidetails}
\end{table}
\subsubsection{Other Datasets}
\label{sec:datasets}

We briefly describe all the remaining datasets used in our experiments. (The relevant primary or secondary language is mentioned in parenthesis, alongside the name of the datasets.) 

\paragraph{IITB Hindi Monolingual Corpus}
(Primary language: \textbf{Hindi})
We extracted 93,000 sentences from the IITB Hindi monolingual corpus%
\footnote{\url{http://www.cfilt.iitb.ac.in/iitb_parallel/iitb_corpus_download/monolingual.hi.tgz}}
, where each sentence has between 4 and 25 tokens. These sentences were used for  unsupervised pretraining.

\paragraph{IITB Parallel Corpus}
(Primary language: \textbf{Hindi})
\label{iitbpar}
We selected 100,000 English-Hindi sentence pairs from IITB parallel corpus~\cite{kunchukuttan2018iit} where the number of tokens in the sentence was greater than 10 for both languages. We used this dataset to further fine-tune the weights of the encoder and decoder layers after unsupervised pretraining.

\paragraph{DuReader~\cite{he2017dureader} Chinese Dataset:}
(Primary language: \textbf{Chinese})
This dataset consists of question-answer pairs along with the question type. We preprocessed and used ``DESCRIPTION" type questions for our experiments, resulting in a total of 8000 instances. From this subset, we created a 6000/1000/1000 split to construct train, development and test sets for our experiments. We also preprocessed and randomly extracted 100,000 descriptions to be used as a Chinese monolingual corpus for the unsupervised pretraining stage.

\paragraph{News Commentary Dataset:}
(Primary language: \textbf{Chinese})
This is a parallel corpus of news commentaries provided by WMT.%
\footnote{\url{http://opus.nlpl.eu/News-Commentary-v11.php}}
It contains roughly 91000 English sentences along with their Chinese translations. We preprocessed this dataset and used this parallel data for fine-tuning the weights of the encoder and decoder layers after unsupervised pretraining.

\paragraph{SQuAD Dataset:}
(Secondary language: \textbf{English})
\label{squad}
This is a very popular English question answering dataset ~\cite{rajpurkar2016squad}.
We used the train split of the pre-processed QG data released by ~\citet{du2017learning} for supervised QG training. This dataset consists of 70,484 sentence-question pairs in English. 

\begin{table*}[htp]
\begin{center}
\small
\scalebox{0.8}{
  \begin{tabular}{|c| l | c | c |c | c | c | c | }
     \hline
     Language & Model& BLEU-1 & BLEU-2 & BLEU-3 & BLEU-4 &METEOR &ROUGE-L \\ 
     \hline
     \multirow{ 2}{*}{Hindi }& Transformer & 28.414 & 18.493 & 12.356 & 8.644 & 23.803 & 29.893\\
     & Transformer+pretraining & 41.059 & 29.294 & 21.403 & 16.047 & 28.159 & 39.395\\
     & \clqg & 41.034 & 29.792 & 22.038 & 16.598 & 27.581 & 39.852\\ 
     & \clqg+parallel & \textbf{42.281} & \textbf{32.074} & \textbf{25.182} & \textbf{20.242} & \textbf{29.143} & \textbf{40.643}\\ 
      \hline
       \multirow{ 2}{*}{Chinese}  & Transformer & 25.52 & 9.22 & 5.14 & 3.25 & 7.64 & 27.40\\
    & Transformer+pretraining &30.38  &14.01 &8.37 &5.18 & \textbf{10.46}& \textbf{32.71}\\
     & \clqg & \textbf{30.69} & \textbf{14.51} & \textbf{8.82} & 5.39 & 10.44 & 31.82\\ 
     & \clqg+parallel & 30.30 & 13.93 & 8.43 & \textbf{5.51} & 10.26 & 31.58\\ 
      \hline
      
  \end{tabular}}
\end{center}
\caption{BLEU, METEOR and ROUGE-L scores on the test set for Hindi and Chinese question generation. Best results for each metric (column) are highlighted in \textbf{bold}.}
\label{tab:main_results}
\end{table*}

\subsection{Implementation Details}
\label{imp_details}
We implemented our model in TensorFlow.\footnote{Code available at https://github.com/vishwajeet93/clqg} We used 300 hidden units for each layer of the transformer with the number of attention heads set to 6. We set the size of BPE embeddings to 300. Our best model uses two independent encoder and decoder layers for both languages, and two shared encoder and decoder layers each. We used a residual dropout set to 0.2 to prevent overfitting. During both the unsupervised pretraining and supervised QG training stages, we used the Adam optimizer ~\cite{kingma2014adam} with a learning rate of $1e{-5}$ and batch size of 64.

\subsubsection{Unsupervised Pretraining}
For Hindi as the primary language, we use 93000 Hindi sentences from the IITB Hindi Monolingual Corpus and around 70000 English sentences from the preprocessed SQuAD dataset for unsupervised pretraining. We pretrain the denoising autoencoders over 15 epochs. For Chinese, we use 100000 Chinese sentences from the DuReader dataset for this stage of training. 

\subsubsection{Supervised Question Generation Training}
We used 73000 sentence-question pairs from SQuAD and 4000 sentence-question pairs from HiQuAD (described in Section ~\ref{hiprep}) to train the supervised QG model in Hindi. We used 6000 Chinese sentence-question pairs from the DuReader dataset to train the supervised QG model in Chinese. 
We initialize all the weights, including the BPE embeddings, from the pretraining phase and fine-tune them until convergence. 
\subsection{Evaluation Methods}
We evaluate our systems and report results on widely used BLEU \cite{papineni2002bleu}, ROUGE-L and METEOR metrics. We also performed a human evaluation study to evaluate the quality of the questions generated. Following ~\citet{kumar2018automating}, we measure the quality of questions in terms of syntactic correctness, semantic correctness and relevance. Syntactic correctness measures the grammatical correctness of a generated question,
semantic correctness measures naturalness of the question, and relevance measures how relevant the question is to the text and answerability of the question from the sentence. 
\section{Results}
\label{res}

We present our automatic evaluation results  in Table~\ref{tab:main_results}, where the primary language is Hindi or Chinese and the secondary language in either setting is English.
We do not report on Chinese as a secondary language owing to the relatively poor quality of the Chinese dataset. 
Here are all the models we compare and evaluate:

\begin{asparaitem}
\item \textbf{Transformer}: We train a transformer model~\cite{vaswani2017attention} using the QG dataset in the primary language. This serves as a natural baseline for comparison.%
\footnote{We also trained a sequence-to-sequence model by augmenting HiQuAD with SQuAD sentences translated into Hindi using Google Translate. This did not perform well giving a BLEU-4 score of 7.54.}
This model consists of a two-layer encoder and a two-layer decoder.%

\item \textbf{Transformer+pretraining}: The above-mentioned \textbf{Transformer} model undergoes an additional step of pretraining. The encoder and decoder layers are pretrained using monolingual data from the primary language. This model will help further demonstrate the value of cross-lingual training.

\item \textbf{\clqg}: This is our main cross-lingual question generation model (described in Section~\ref{sec:main_model}) where the encoder and decoder layers are initialized in an unsupervised pretraining phase using primary and secondary language monolingual corpora, followed by a joint supervised QG training using QG datasets in the primary and secondary languages.

\item \textbf{\clqg+parallel}: The \clqg\xspace model undergoes further training using a parallel corpus (with primary language as source and secondary language as target). After unsupervised pretraining, the encoder and decoder weights are fine-tuned using the parallel corpus. This fine-tuning further refines the language models for both languages and helps enforce the shared latent space across both languages. 

\par We observe in Table~\ref{tab:main_results} that  \textbf{\clqg+parallel} outperforms all the other models for Hindi. For Chinese, parallel fine-tuning does not give significant improvements over \clqg; this could be attributed to the parallel corpus being smaller in size (when compared to Hindi) and domain-specific (i.e. the news domain). 
\end{asparaitem}

\begin{table}[b!]
\begin{center}
\small
\scalebox{0.8}{
  \begin{tabular}{| l | c | c | c | c |c | c | }
    \hline
     \multirow{2}{*}{Model}& \multicolumn{2}{c|}{Syntax} & \multicolumn{2}{c|}{Semantics} & \multicolumn{2}{c|}{Relevance} \\ \cline{2-7}
            & Score & Kappa & Score & Kappa & Score & Kappa \\
    \hline
     \textbf{Transformer} & 71 & 0.239 & 62.5 & 0.46 & 32 & 0.75  \\
     \hline
     \textbf{\clqg}& \textbf{72} & 0.62 & \textbf{68.5} & 0.82 & \textbf{54} & 0.42 \\
     \textbf{+parallel} & & & & & &\\
     \hline
  \end{tabular}}
\end{center}
\caption{Human evaluation results as well as inter-rater agreement (column ``Kappa'') for each model on the Hindi test set. The scores are between 0-100, 0 being the worst and 100 being the best. Best results for each metric (column) are in \textbf{bold}. The three evaluation criteria are: (1) syntactic correctness ({Syntax}), (2) semantic correctness ({Semantics}), and (3) relevance to the paragraph ({Relevance}).}
\label{heresults}
\end{table}

\begin{figure*}[t]
\centering
\begin{subfigure}{0.32\textwidth}
\includegraphics[width=0.98\linewidth]{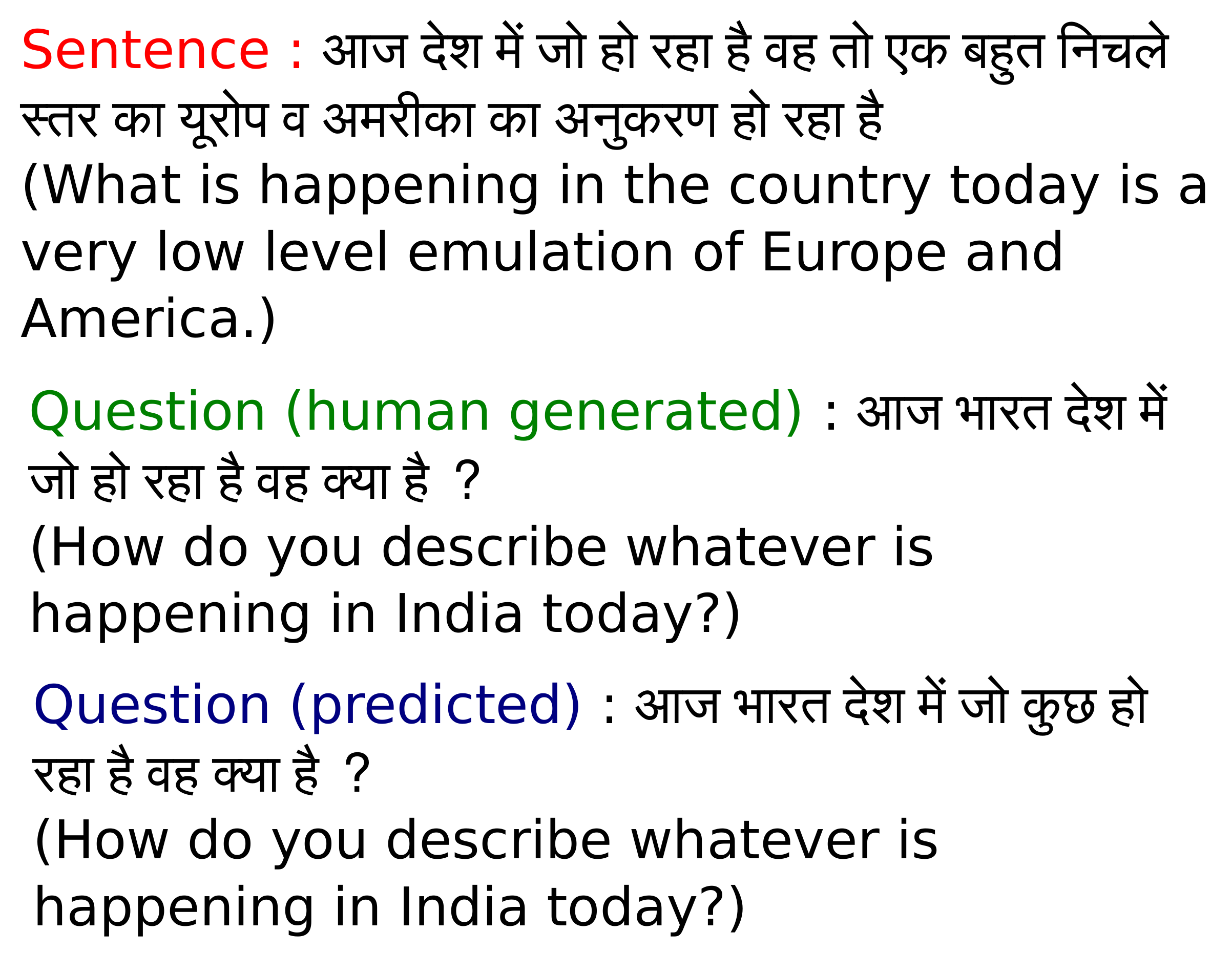} 
\caption{}
\label{fig:correct-1}
\end{subfigure}
\begin{subfigure}{0.32\textwidth}
\includegraphics[width=0.98\linewidth]{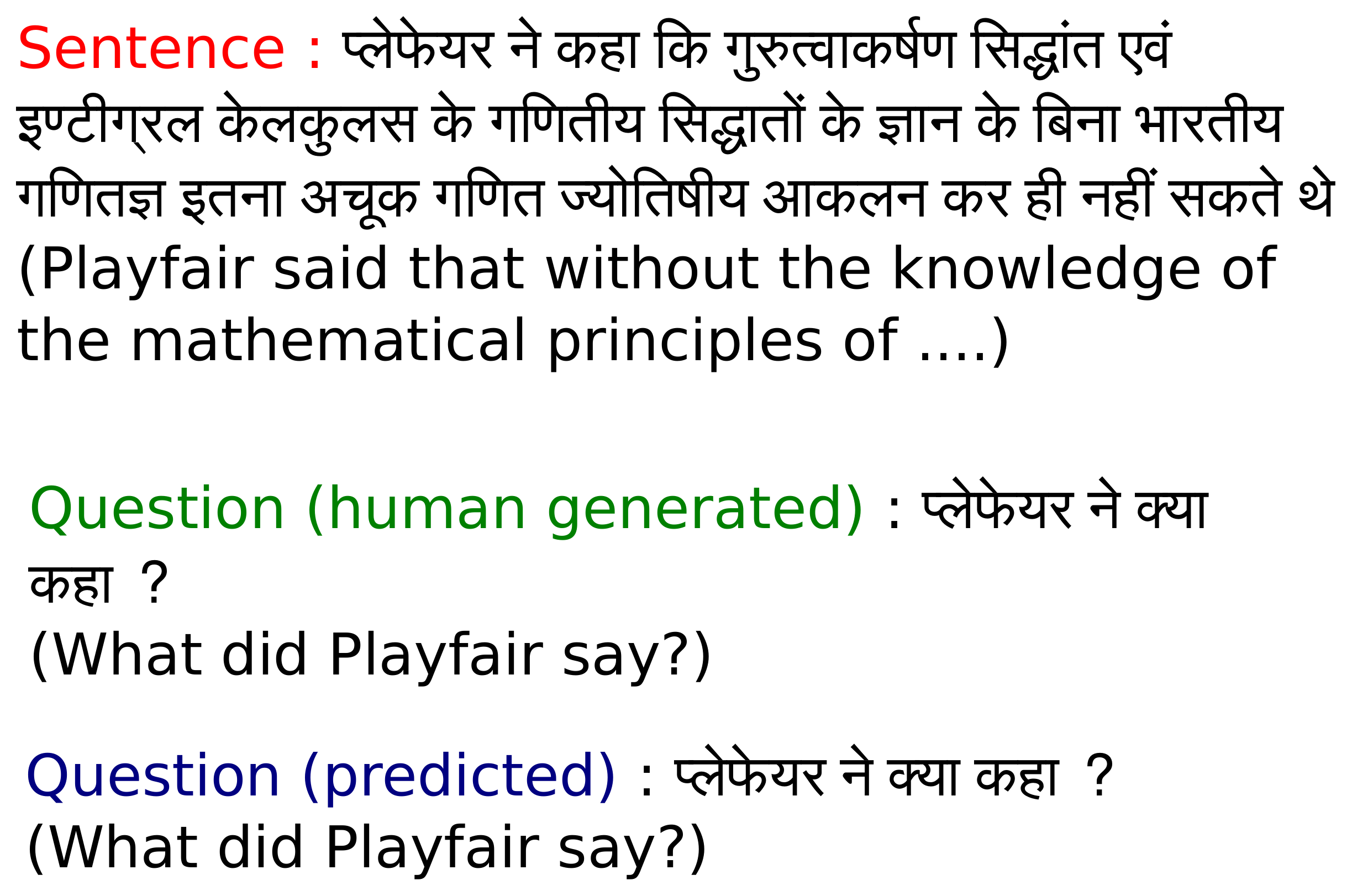}
\caption{}
\label{fig:correct-2}
\end{subfigure}
\begin{subfigure}{0.32\textwidth}
\includegraphics[width=0.98\linewidth]{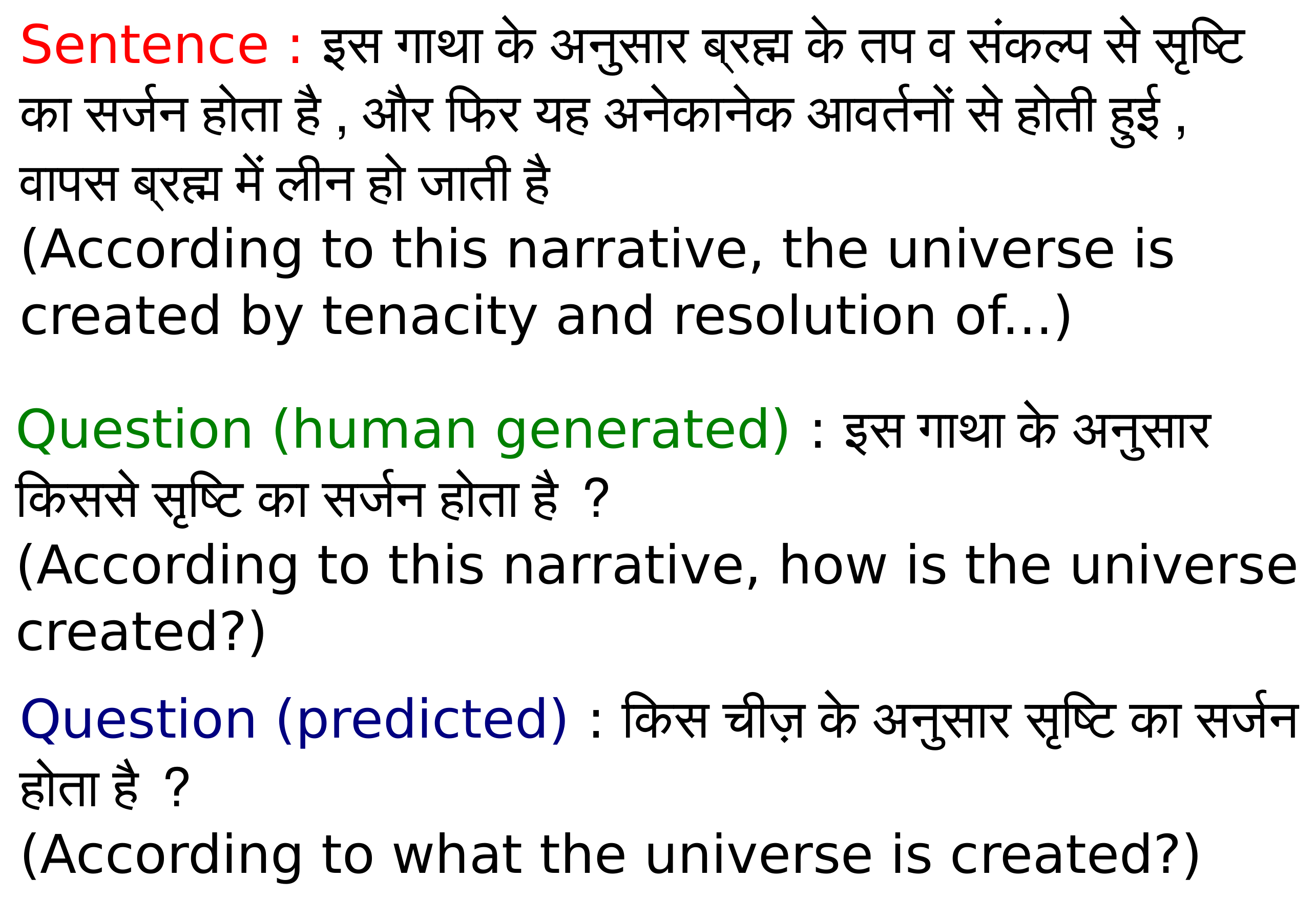}
\caption{}
\label{fig:correct-3}
\end{subfigure}
\caption{Three examples of correctly generated Hindi questions by our model, further analyzed in Section~\ref{ssec:error}.} 
\label{fig:correct}
\end{figure*}

\begin{figure}[t]
\begin{subfigure}{0.36\textwidth}
\centering
\includegraphics[width=\linewidth]{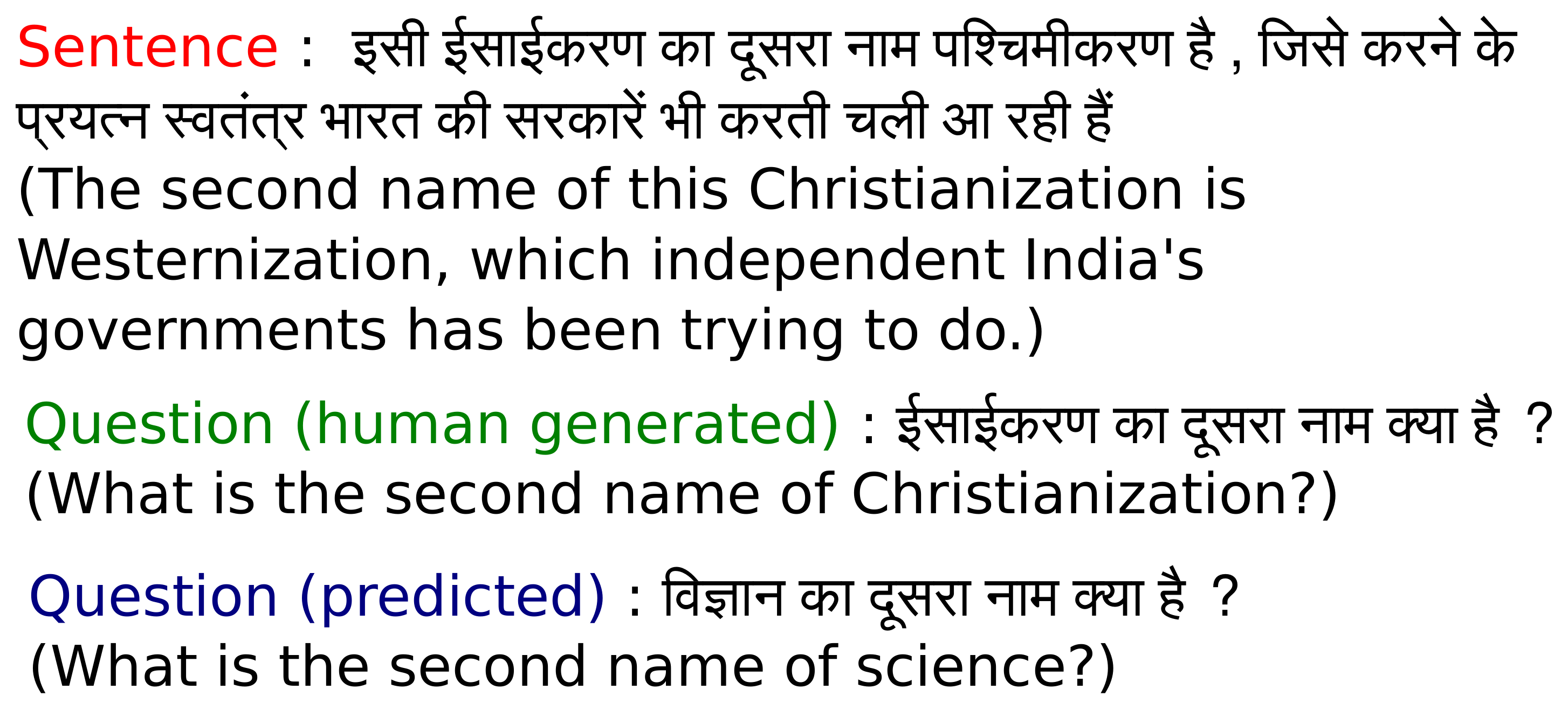}
\caption{}
\label{fig:incorrect-1}
\end{subfigure}
\begin{subfigure}{0.4\textwidth}
\centering
\includegraphics[width=\linewidth]{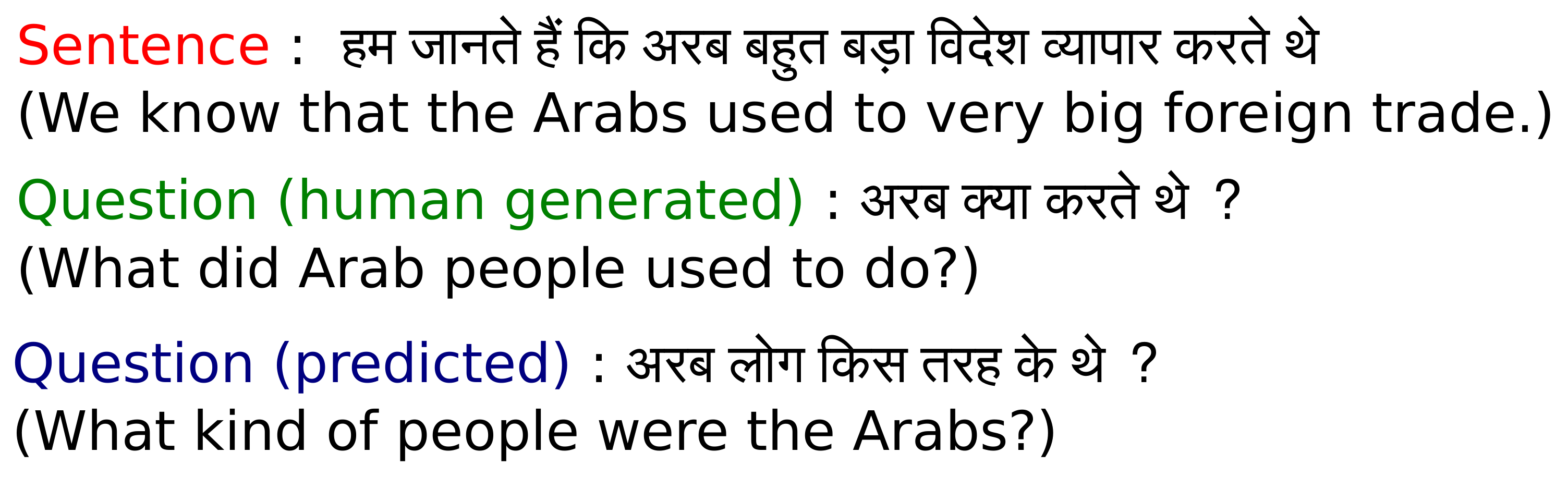}
\caption{}
\label{fig:incorrect-2}
\end{subfigure}
\caption{Two examples of incorrectly generated Hindi questions by our model, further analyzed in Section~\ref{ssec:error}.}
\label{fig:incorrect}
\end{figure}
\section{Discussion and Analysis}
\label{analysis}

\begin{figure}[t]
\begin{subfigure}{0.36\textwidth}
\centering
\includegraphics[width=\linewidth]{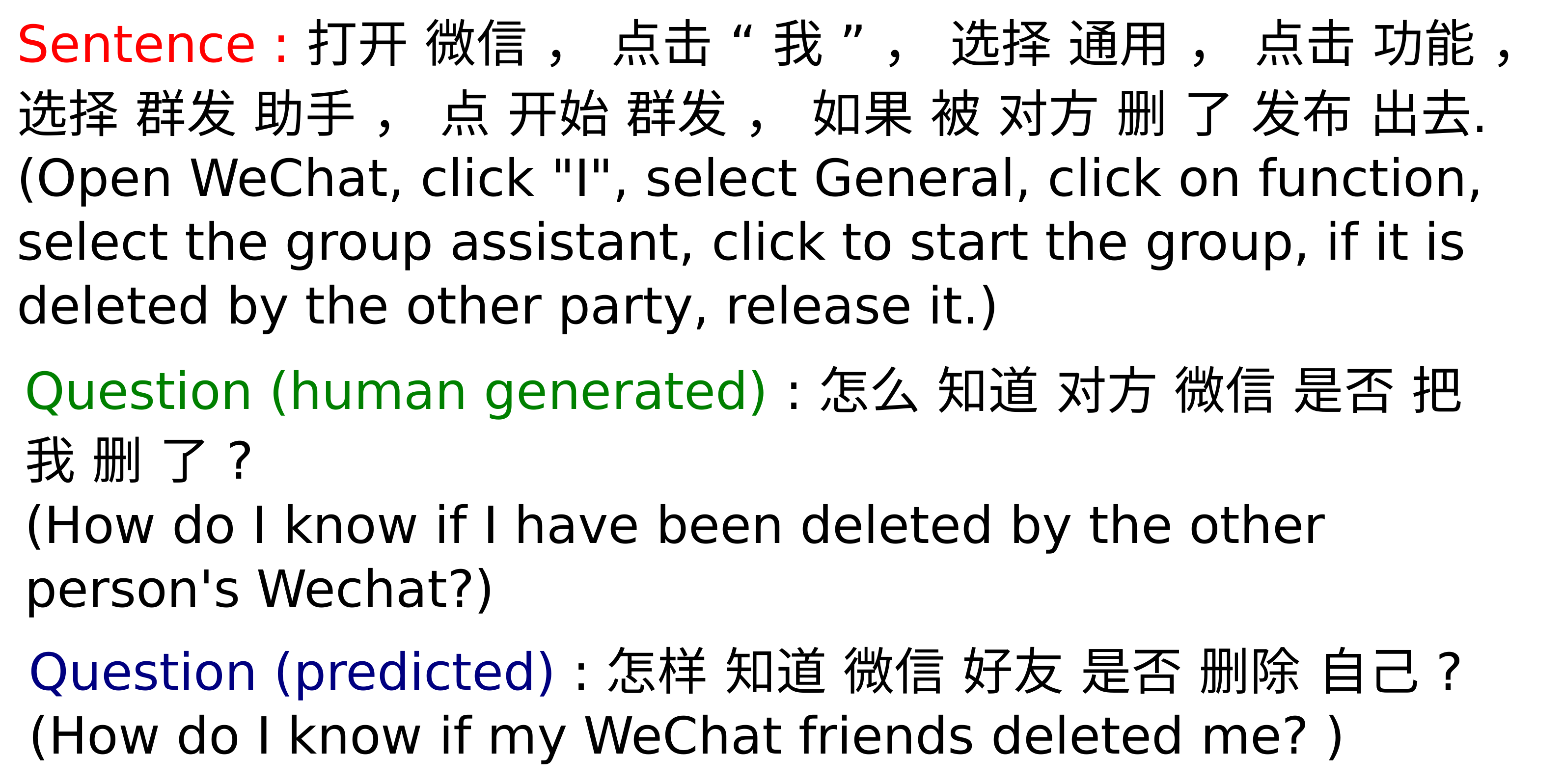}
\caption{}
\label{fig:incorrect-1-ch}
\end{subfigure}
\begin{subfigure}{0.4\textwidth}
\centering
\includegraphics[width=\linewidth]{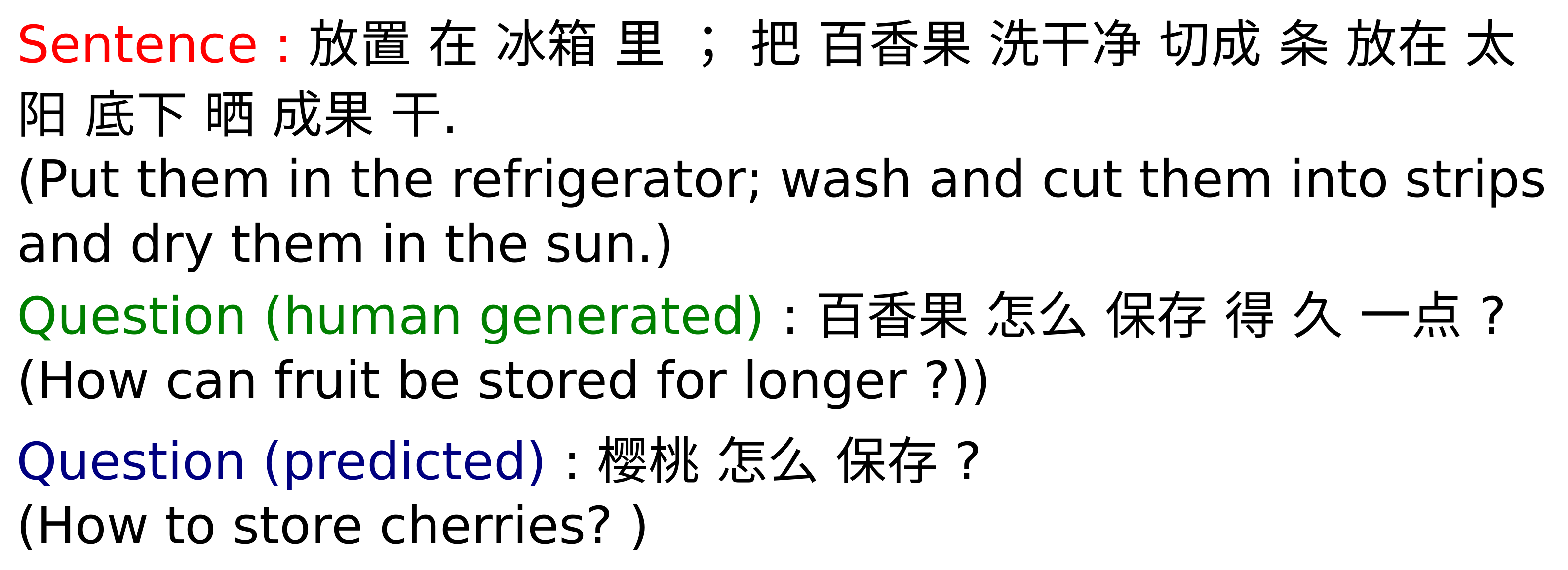}
\caption{}
\label{fig:incorrect-2-ch}
\end{subfigure}
\caption{Automatic QG from Chinese text.}
\label{fig:ch_example}
\end{figure}

We closely inspect our cross-lingual training paradigm using (i) a human evaluation study in Section~\ref{ssec:humaneval} (ii) detailed error analysis in Section~\ref{ssec:error} and (iii) ablation studies in Section~\ref{ablation}. All the models analyzed in this section used Hindi as the primary language.%
\footnote{Figure~\ref{fig:ch_example} shows two examples of correctly generated Chinese questions.}

\subsection{Human evaluation}
\label{ssec:humaneval}
We conduct a human evaluation study comparing the questions generated by the \textbf{Transformer} and \textbf{\clqg +parallel} models. We randomly selected a  subset of 100 sentences from the Hindi test set and generated questions using both models. We presented these sentence-question pairs for each model to three language experts and asked for a binary response on three quality parameters namely syntactic correctness, semantic correctness and relevance. The responses from all the experts for each parameter was averaged for each model to get the final numbers shown in Table~\ref{heresults}. Although we perform comparably to the baseline model on syntactic correctness scores, we obtain significantly higher agreement across annotators using our cross-lingual model. Our cross-lingual model performs significantly better than the \textbf{Transformer} model on ``Relevance" at the cost of agreement. On semantic correctness, we perform signficantly better both in terms of the score and agreement statistics.

\begin{table*}[htp]
\begin{center}
\small
\scalebox{0.8}{
  \begin{tabular}{| l | c | c |c | c | c | c | }
    \hline
     Model& BLEU-1 & BLEU-2 & BLEU-3 & BLEU-4 &METEOR &ROUGE-L \\ \hline
     \clqg\xspace (no pretraining) & 31.707 & 20.727 & 13.954 & 9.862 & 24.209 & 32.332\\
     \hline
     \clqg & 41.034 & 29.792 & 22.038  & 16.598 & 27.581 & 39.852\\ 
      \hline
    \clqg + parallel & \textbf{42.281} & \textbf{32.074} & \textbf{25.182} & \textbf{20.242} & \textbf{29.143} & \textbf{40.643}\\ 
      \hline
  \end{tabular}}
\end{center}
\caption{Ablation study showing the importance of both unsupervised and unsupervised pretraining for Hindi}
\label{tab:pretrain}
\end{table*}

\label{sssec:sec_lang}
\begin{table*}[!htp]
\begin{center}
\small
\scalebox{0.8}{
  \begin{tabular}{| l | c | c |c | c | c | c | }
    \hline
     Dataset & BLEU-1 & BLEU-2 & BLEU-3 & BLEU-4 &METEOR &ROUGE-L \\ 
     \hline
     Hindi QG only & 41.66  & 31.576 & 24.572 & 19.538 & 28.665 & \textbf{40.765}\\
     \hline
     Hindi QG + English QG & \textbf{42.281} & \textbf{32.074} & \textbf{25.182} & \textbf{20.242} & \textbf{29.143} & 40.643\\ 
      \hline
  \end{tabular}}
\end{center}
\caption{Ablation study showing the importance of using English QG data for Hindi QG}
\label{tab:sec_resources}
\end{table*}

\subsection{Error Analysis}
\label{ssec:error}
\paragraph{Correct examples: }We show several examples where our model is able to generate semantically and syntactically correct questions in Figure \ref{fig:correct}. Figure \ref{fig:correct-2} shows our model is able to generate questions that are identical to  human-generated questions. Fig.~\ref{fig:correct-3} demonstrates that our model can generate new questions which clearly differ from the human-generated questions but are syntactically correct, semantically correct and relevant to the text. Fig.~\ref{fig:correct-1} shows a third question  which differs from the human-generated question in only a single word but does not alter its quality.

\paragraph{Incorrect examples: } We also present a couple of examples where our model is unable to generate good questions and analyze possible reasons for the same. In Fig.~\ref{fig:incorrect-1}, the model captures the type of question correctly but gets the main subject of the sentence wrong. On the other hand, Fig.~\ref{fig:incorrect-2} shows a question which is syntactically correct and relevant to the main subject, but is not consistent with the given sentence.

\subsection{Ablation Studies}
\label{ablation}
We performed two experiments to better understand the role of each component in our model towards automatic QG from Hindi text. 
\subsubsection{Importance of unsupervised pretraining}
\label{sssec:wo_pre}

We construct a model which does not employ any unsupervised or supervised pretraining but uses the same network architecture. This helps in studying the importance of pretraining in our model.
We present our results in Table~\ref{tab:pretrain}. We observe that our shared architecture does not directly benefit from the English QG dataset with simple weight sharing. Unsupervised pretraining (with back-translation) helps the shared encoder and decoder layers capture higher-level language-independent information giving an improvement of approximately 7 in BLEU-4 scores. Additionally, the use of parallel data for fine-tuning unsupervised pretraining aids this process further by improving BLEU-4 scores by around 3 points. 

\subsubsection{Importance of secondary language resources}

To demonstrate the improvement in Hindi QG from the relatively larger English SQuAD dataset, we show results of using only HiQuAD during the main task in Table~\ref{tab:sec_resources}; unsupervised and supervised pretraining are still employed. We obtain modest performance improvements on the standard evaluation metrics (except ROUGE-L) by using English SQuAD data in the main task. These improvements (albeit small) demonstrate that our proposed cross-lingual framework is a step in the right direction towards leveraging information from a secondary language.

\begin{figure}
    \centering
    \includegraphics[width=\linewidth]{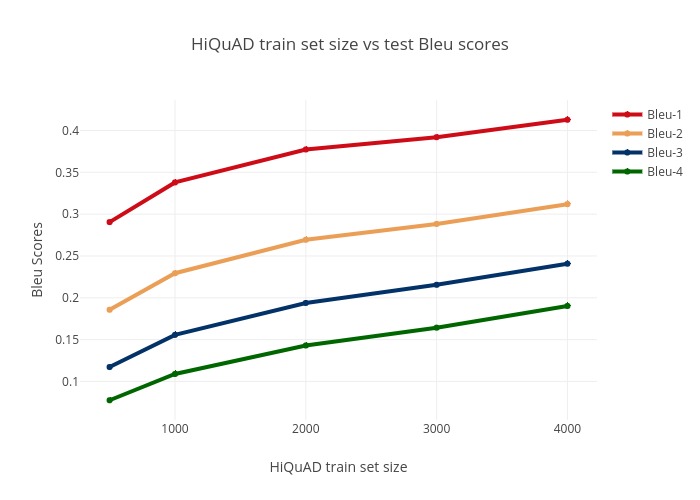}
    \caption{Trade-off between HiQuAD training dataset size and BLEU scores.}
    \label{fig:ds_vs_bleu}
\end{figure}

\subsection{How many sentence-question pairs are needed in the primary language?}
To gain more insight into how much data is required to be able to generate questions of high quality, Fig.~\ref{fig:ds_vs_bleu} presents a plot of BLEU scores when the number of Hindi sentence-question pairs is varied. Here, both unsupervised and supervised pretraining are employed but the English SQuAD dataset is not used. After significant jumps in BLEU-4 performance using the first 2000 sentences, we see a smaller but steady improvement in performance with the next set of 2000 sentences.

\section{Conclusion}
\label{conc}
Neural models for automatic question generation using the standard sequence to sequence paradigm have been shown to perform reasonably well for languages such as English, which have a large number of training instances. However, large training sets are not available for most languages. To address this problem, we present a cross-lingual model that leverages a large QG dataset in a secondary language (along with monolingual data and parallel data) to improve QG performance on a primary language with a limited number of QG training pairs. In future work, we will explore the use of cross-lingual embeddings to further improve performance on this task.

\section{Acknowledgments}
The authors thank the anonymous reviewers for their insightful comments that helped improve this paper. The authors also gratefully acknowledge support from IBM Research, India (specifically the IBM AI Horizon Networks - IIT Bombay initiative). 

\bibliography{acl2019}
\bibliographystyle{acl_natbib}

\appendix

\end{document}